\documentclass[10pt,letterpaper,twocolumn]{article}
\usepackage[margin=0.75in,columnsep=0.28in]{geometry}
\usepackage[T1]{fontenc}
\usepackage{newtxtext}
\usepackage[hyphens]{url}
\usepackage{graphicx}
\usepackage[round,authoryear]{natbib}
\usepackage{caption}
\usepackage{amsmath}
\usepackage{amssymb}
\usepackage{algorithm}
\usepackage{algorithmic}
\usepackage{newfloat}
\usepackage{listings}
\usepackage{booktabs}
\usepackage{xcolor}
\usepackage{microtype}

\DeclareCaptionStyle{ruled}{labelfont=normalfont,labelsep=colon,strut=off}
\floatstyle{ruled}
\newfloat{listing}{tb}{lst}{}
\floatname{listing}{Listing}

\title{\textbf{TraceCLIP: Recovering Local Semantics from Patch-to-CLS Contributions}}
\author{%
\begin{tabular}{c}
Xinran Liu,
Shouqian Shi\textsuperscript{\dag},
Yutong Chen,\quad
Ge Wang,
Xin-Wei Yao,
Sheng Zhong\\[0.35em]
\small \textsuperscript{\dag}Corresponding author: \texttt{sqlite@nju.edu.cn}
\end{tabular}%
}
\date{}

\begin{document}

\maketitle
\begin{abstract}
Dense vision-language understanding, including object localization, region
recognition, and open-vocabulary semantic segmentation, requires associating
language concepts with spatially grounded visual regions. CLIP provides a
strong foundation for these tasks by learning a shared image-text embedding
space from large-scale contrastive pre-training. However, its image-level
objective aligns text with a CLS-derived global representation, leaving local
vision-language correspondence only indirectly constrained. Existing methods
either introduce additional supervision, external models, or task-specific
adaptation, while training-free approaches mainly recover dense responses
from existing patch features without examining where local semantics become
most accessible within CLIP. We introduce \textbf{TraceCLIP}, a training-free
framework that recovers latent patch-level semantic evidence by isolating the
patch-specific terms written into the CLS attention output. TraceCLIP further
converts contribution-derived semantic responses into a semantic-geodesic
topology gate that calibrates final-layer patch affinity for dense feature
reconstruction. Diagnostic experiments show that these contribution features exhibit strong
local semantic discrimination and text-conditioned spatial alignment. On eight zero-shot semantic segmentation benchmarks,
TraceCLIP achieves gains of 1.3--4.5 points in average mIoU over the
strongest prior training-free methods across both backbones and background
settings, without additional training, external vision foundation models, or
region-level supervision. More broadly, these findings suggest that
spatially localized semantics may remain accessible within the internal
construction of globally aligned representations.
\end{abstract}

\section{Introduction}

Dense vision-language understanding requires local representations that are
both spatially grounded and aligned with language. Such representations
support open-vocabulary object detection, phrase grounding, region
recognition, and semantic segmentation
\cite{minderer2022simple,li2022grounded,liu2024grounding,
zhong2022regionclip,li2022language,rao2022denseclip}. CLIP provides a strong
foundation for these tasks by learning a joint image-text embedding space
from large-scale contrastive pre-training \cite{radford2021learning}.
However, in ViT-based CLIP encoders \cite{dosovitskiy2021image}, the image
representation used for contrastive alignment is derived from the final CLS
token. This produces strong global image-text alignment, but leaves
patch-level vision-language correspondence only indirectly constrained.

Existing approaches transfer CLIP's global alignment to local prediction
through region-level supervision, mask annotations, image-level dense
adaptation, or task-specific training
\cite{gu2022open,zhong2022regionclip,li2022language,xu2022groupvit,
ghiasi2022scaling,zhou2022extract,liang2023open}. Although effective, these
methods introduce additional optimization, supervision, or external
components. Training-free methods instead keep CLIP frozen and improve dense
inference through attention redesign, spatial consistency, or affinity-based
feature propagation
\cite{wang2024sclip,lan2024clearclip,hajimiri2025pay}. Their primary focus is
how to transform existing patch states into better dense responses. A more
fundamental question remains less explored: where within CLIP's computation
does local semantic evidence become most directly accessible?

This question extends beyond improving a particular dense prediction
pipeline. It reflects a broader representation problem in globally aligned
models: information useful for local or structured prediction may participate
in constructing the global representation. Recovering such information from
internal computation provides a complementary direction to additional
training or architectural adaptation.

We revisit the construction of CLIP's language-aligned image representation.
Within each self-attention block, the CLS attention update is an additive
combination of weighted value components written from individual token
positions. The patch-specific terms in this update therefore provide a
direct trace of how local visual evidence contributes to the CLS pathway.
By isolating these patch-to-CLS contributions and reading them out in the
CLIP joint embedding space, we obtain patch-level features that can be
directly compared with text embeddings. Rather than treating decomposition
only as an interpretability tool, we use it as a representation-recovery
mechanism that turns the internal terms constructing a global embedding into
reusable local representations.

Our diagnostic analysis supports this view. Across region classification,
text-conditioned localization, and late-block component analysis,
patch-to-CLS contribution features are substantially more discriminative
and spatially aligned than final patch-token states. The results further
show that local semantic evidence is clearly exposed in the attention
contribution pathway but becomes less accessible after residual and MLP
mixing. These findings indicate that frozen CLIP already contains useful
spatially grounded semantic evidence, provided that it is read from the
appropriate internal pathway.

Based on this observation, we propose \textbf{TraceCLIP}, a training-free
framework for recovering dense CLIP features from patch-to-CLS
contributions. Direct contribution--text matching provides informative
local semantic responses, but independent patch-wise predictions lack the
spatial coherence required for dense segmentation. TraceCLIP therefore uses
these responses to construct preliminary semantic labels and confidence
scores, from which it derives a semantic-geodesic topology gate. The gate
calibrates final-layer query-query affinity before value aggregation,
allowing dense reconstruction to preserve visual correlation while
suppressing propagation across confident semantic boundaries. The entire
framework keeps both CLIP encoders frozen and requires no additional
training, external vision foundation models, or region-level supervision.

Our contributions are summarized as follows:
\begin{itemize}
\item We identify a representation gap between final patch states and the
patch-specific terms that construct CLIP's global embedding, showing that
the patch-to-CLS contribution pathway provides an explicit internal source
of localized vision-language evidence.

\item We provide diagnostic evidence through region classification,
text-conditioned localization, and late-block component analysis, showing
that contribution features are more discriminative and spatially grounded
than standard patch-level readouts.

\item We propose \textbf{TraceCLIP}, which converts contribution-derived
semantic responses into a semantic-geodesic topology gate for calibrating
final-layer patch affinity and reconstructing dense features without
training. Experiments on eight zero-shot semantic segmentation benchmarks show
that TraceCLIP improves average mIoU over the strongest prior training-free
methods by 1.3--4.5 points across both evaluated backbones and background
settings.
\end{itemize}
\section{Related Work}

\subsection{Global-to-Local Transfer in CLIP}

CLIP learns a joint image-text embedding space from image-level contrastive supervision and has become a strong foundation for open-vocabulary visual recognition \cite{radford2021learning}. However, spatially grounded vision-language prediction requires localized representations that can be compared with text embeddings at the object, region, mask, or pixel level. A broad body of work therefore studies how to transfer CLIP's global image-text alignment to local prediction. Representative methods extend CLIP-style vision-language knowledge to open-vocabulary detection, grounding, region recognition, and semantic segmentation by introducing localization-aware training objectives, additional prediction modules, or task-specific adaptation strategies \cite{gu2022open,zhong2022regionclip,minderer2022simple,li2022grounded,liu2024grounding,li2022language,xu2022groupvit,ghiasi2022scaling,rao2022denseclip,cha2023learning,liang2023open}.

These methods differ in task formulation and supervision granularity, but they commonly introduce extra learning or adaptation to obtain localized vision-language alignment beyond the original CLIP training objective. Our work studies a different setting: the CLIP image encoder remains frozen, and local semantic evidence is recovered from its internal computation rather than learned through additional local supervision or adaptation modules.

\subsection{Training-Free Dense Prediction with Frozen CLIP}

A closely related line of work repurposes frozen CLIP for dense
vision-language inference without parameter updates. MaskCLIP shows that dense
predictions can be extracted from CLIP features in a training-free manner
\cite{zhou2022extract}. ReCo retrieves concept-relevant image collections and
co-segments their shared visual regions to transfer CLIP-derived semantic cues
to zero-shot segmentation~\cite{shin2022reco}. Subsequent methods improve
frozen-CLIP dense prediction by modifying attention computation, adjusting
late visual blocks, enforcing local consistency, introducing proxy spatial
correspondences, or refining dense responses with semantic feedback
\cite{wang2024sclip,lan2024clearclip,hajimiri2025pay}.

These methods demonstrate that frozen CLIP contains useful signals for dense
prediction and that its inference-time computation can be adapted without
retraining. Their focus is mainly on how to obtain better dense responses from
existing patch-level states, attention maps, or affinity structures. However,
instead of focusing on how patch-token features should be transformed, we
shift the emphasis to where the most effective local semantic evidence
resides. We accordingly explore the patch-to-CLS contribution path in detail,
maintaining the training-free style while improving dense prediction accuracy.

\subsection{CLIP Representation Analysis}
In ViT-based CLIP, the image representation used for contrastive learning is derived from the final class token, commonly referred to as the CLS token, while patch tokens interact with it through self-attention \cite{dosovitskiy2021image,radford2021learning}. This architecture suggests that patch-level visual evidence may participate in constructing the global image representation aligned with text.

Recent work analyzes CLIP image representations through text-based decomposition, where the final image embedding is decomposed into layer-, head-, and patch-level contributions and interpreted with text embeddings \cite{gandelsman2024interpreting}. This provides a way to trace how local visual evidence contributes to the CLS-derived image representation. Different from using decomposition mainly for post-hoc interpretation of global image-text similarity, we use the patch-to-CLS contribution path as a source of local vision-language features for dense prediction. We further convert contribution responses into a semantic affinity prior, making decomposition part of a training-free dense feature recovery method.

\section{Patch-to-CLS Contributions as Local Semantic Evidence}

\subsection{Patch-to-CLS Contribution Decomposition}

A ViT-based CLIP image encoder represents an image at layer $l$ as
$X^l=[x_0^l,x_1^l,\ldots,x_N^l]$, where $x_0^l$ is the CLS token and
$x_i^l$, $i\geq1$, denotes a patch token. Because the multi-head
self-attention output is additive over attention heads and source-token
positions, the CLS attention update can be decomposed into token-head
terms. We therefore isolate the image-patch terms written into the CLS
attention output and use them as local vision-language features.

For the multi-head self-attention block at layer $l$, let
$A_{0i}^{l,h}$ denote the attention coefficient from the CLS token to
token $i$ in head $h$, and let $W_{OV}^{l,h}$ denote the corresponding
value-to-output transition matrix. We define the CLIP-space readout of
the contribution of token $i$ through head $h$ to the CLS attention
output as
\begin{equation}
C_i^{l,h}
=
W_{\mathrm{proj}}
\left(
A_{0i}^{l,h}
W_{OV}^{l,h}
\operatorname{LN}(x_i^l)
\right),
\label{eq:token_head_contribution}
\end{equation}
where $\operatorname{LN}$ denotes the pre-attention layer normalization
and $W_{\mathrm{proj}}$ denotes the CLIP visual projection.
Eq.~\ref{eq:token_head_contribution} maps a token-head term in the CLS
attention update into the CLIP joint embedding space.

To obtain a patch-level feature, we retain the image-patch terms and
aggregate their contributions over attention heads. For
$i=1,\ldots,N$, we define
\begin{equation}
C_i^l
=
\sum_{h=1}^{H} C_i^{l,h}.
\label{eq:patch_cls_contribution}
\end{equation}
We refer to $C_i^l$ as the patch-to-CLS contribution feature of patch
$i$ at layer $l$. Since it represents the component written from patch
$i$ into the CLS attention output and is read out in the CLIP joint
embedding space, it can be directly compared with text embeddings and
used as a local vision-language feature.

\begin{figure}[t]
    \centering
    \includegraphics[width=\columnwidth]{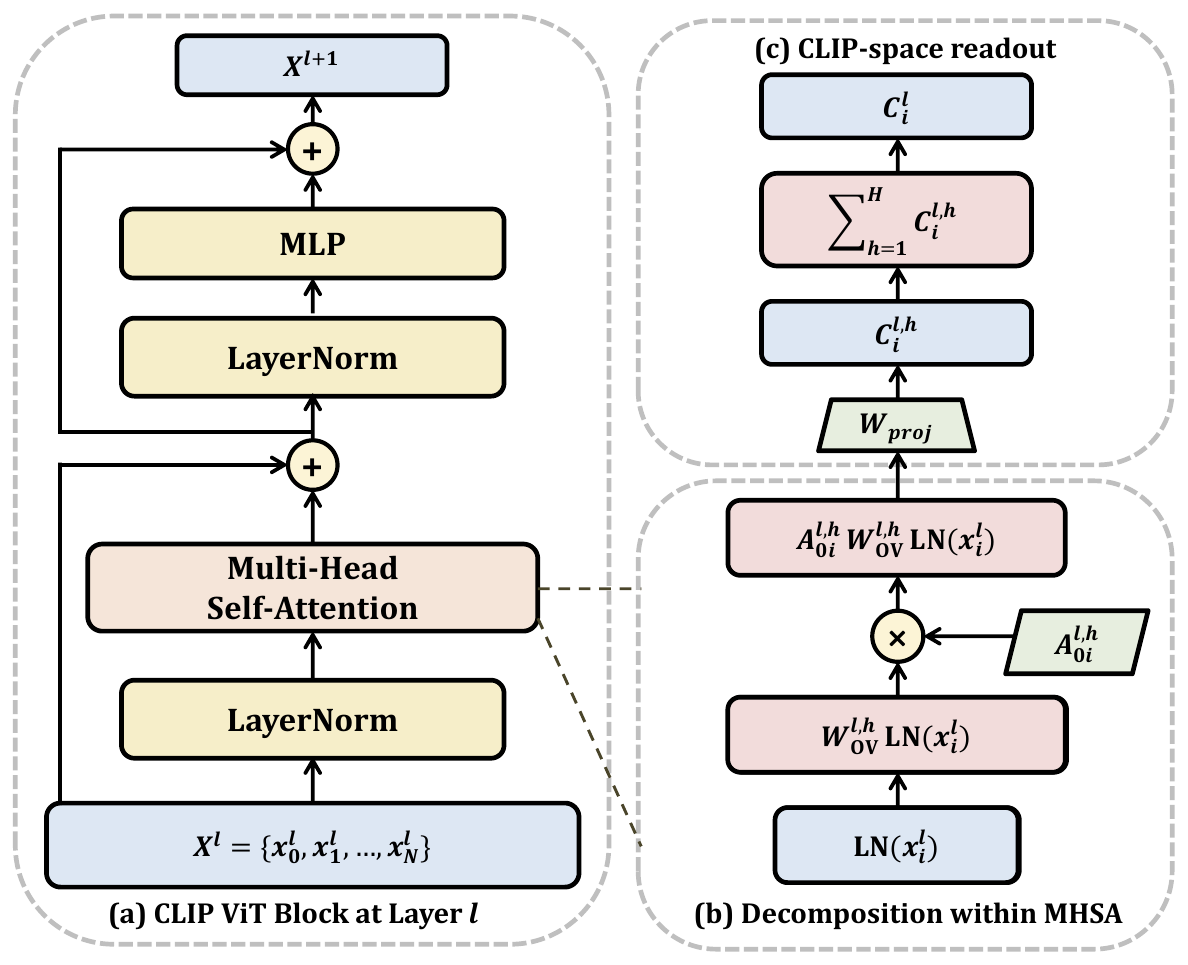}
    \caption{Patch-to-CLS contribution decomposition.
    $A_{0i}^{l,h}$ denotes the attention weight from the CLS token to
    patch $i$ in head $h$, $W_{OV}^{l,h}$ is the corresponding
    value-to-output transition matrix, and $W_{\mathrm{proj}}$ is the
    CLIP visual projection. The resulting CLIP-space token-head
    contributions $C_i^{l,h}$ are aggregated over heads to obtain
    $C_i^l$.}
    \label{fig:contribution_decomposition}
\end{figure}

\subsection{Probing Local Semantics in Contribution Features}

We conduct diagnostic experiments on the COCO validation set
\cite{lin2014microsoft} to answer three questions: whether patch-to-CLS
contributions contain category-discriminative local semantics, whether their
responses are spatially aligned with the corresponding objects, and where
the gap between contribution features and final patch states arises.

\paragraph{Feature sources and probing setup.}
We compare three representative feature sources.
\textbf{1)} \emph{Final patch}, obtained by applying CLIP's visual
post-layer normalization and visual projection to the final patch-token
states.
\textbf{2)} $V^{(-1)}$, the value feature from the last self-attention
block, merged across heads and projected into the CLIP joint embedding
space.
\textbf{3)} $C^{(-k)}$, the patch-to-CLS contribution feature from the
$k$-th visual block from the end. Compared with $V^{(-1)}$,
$C^{(-k)}$ additionally incorporates the attention weight assigned by
the CLS query to each patch and therefore represents the patch-specific
content written into the CLS attention update.

\paragraph{Region-level semantic discrimination.}
We first test whether each feature source supports category discrimination
within a localized region. Given a ground-truth box $b$, we average the
features of patches whose centers fall inside the box and compare the pooled
feature with category text embeddings:
\begin{equation}
r_b=\frac{1}{|\mathcal{P}(b)|}\sum_{i\in \mathcal{P}(b)} f_i,
\quad
s(b,y)=\cos(r_b,t_y),
\label{eq:region_readout}
\end{equation}
where $f_i$ is the feature of patch $i$, $t_y$ is the text embedding of
category $y$, and $\mathcal{P}(b)$ denotes the corresponding patch set.
Ground-truth boxes isolate representation quality from proposal generation
and non-maximum suppression. We report the Top-1 category accuracy as
\emph{Region Acc.}

\paragraph{Text-conditioned spatial alignment.}
Region classification does not determine whether responses are spatially
concentrated on the target object. We therefore compare each patch feature
with the target-category text embedding and report two localization metrics.
\emph{Pointing Acc.} measures whether the highest-response patch lies inside
the ground-truth box. \emph{In--Out Gap} is the mean target-category response
inside the box minus that outside the box, averaged over all evaluated
objects. The two metrics respectively measure peak localization and overall
object-background separation.

\begin{table}[!t]
\centering
\small
\begin{tabular}{lccc}
\toprule
\textbf{Feature}
& \textbf{Reg. Acc.}
& \textbf{Point. Acc.}
& \textbf{In--Out Gap} \\
\midrule

\multicolumn{4}{c}{\textit{ViT-B/16}} \\
\midrule
Final patch
& 11.3
& 19.9
& -0.042 \\

$V^{(-1)}$
& 8.4
& 21.8
& 0.000 \\

$C^{(-4)}$
& 7.9
& 38.4
& 0.021 \\

$C^{(-3)}$
& 25.5
& 60.9
& 0.132 \\

$C^{(-2)}$
& 37.0
& 62.7
& 0.215 \\

$C^{(-1)}$
& \textbf{43.1}
& \textbf{65.7}
& \textbf{0.275} \\

\midrule
\multicolumn{4}{c}{\textit{ViT-L/14}} \\
\midrule
Final patch
& 8.7
& 14.0
& -0.048 \\

$V^{(-1)}$
& 7.4
& 26.4
& 0.005 \\

$C^{(-4)}$
& 35.8
& 60.8
& 0.188 \\

$C^{(-3)}$
& 32.9
& 60.9
& 0.180 \\

$C^{(-2)}$
& 42.8
& 64.9
& 0.257 \\

$C^{(-1)}$
& \textbf{47.1}
& \textbf{68.9}
& \textbf{0.287} \\

\bottomrule
\end{tabular}

\caption{Local semantic probing on the COCO validation set.
Reg. Acc. and Point. Acc. are reported in percentages.
In--Out Gap denotes the inside--outside target-category response
difference.}
\label{tab:local_semantic_probing}
\end{table}

Table~\ref{tab:local_semantic_probing} reveals a clear accessibility gap
between contribution features and conventional patch-level readouts.
Final patch states and $V^{(-1)}$ provide weak region discrimination and
localization, whereas near-final contribution features substantially improve all three
metrics. For ViT-B/16, $C^{(-1)}$ raises Region Acc. and Pointing Acc. from
11.3\% and 19.9\% to 43.1\% and 65.7\%, respectively. For ViT-L/14, the
corresponding improvements are from 8.7\% and 14.0\% to 47.1\% and 68.9\%.

Performance generally improves toward the final visual blocks, with
$C^{(-1)}$ achieving the best result on every metric and both backbones.
The gap between $V^{(-1)}$ and $C^{(-1)}$ suggests that CLS attention
weighting selects patch evidence that is more strongly aligned with the
language-supervised global pathway. The positive In--Out Gap further shows
that this evidence is spatially concentrated within target object regions.

\begin{figure}[t]
    \centering
    \includegraphics[width=0.95\columnwidth]{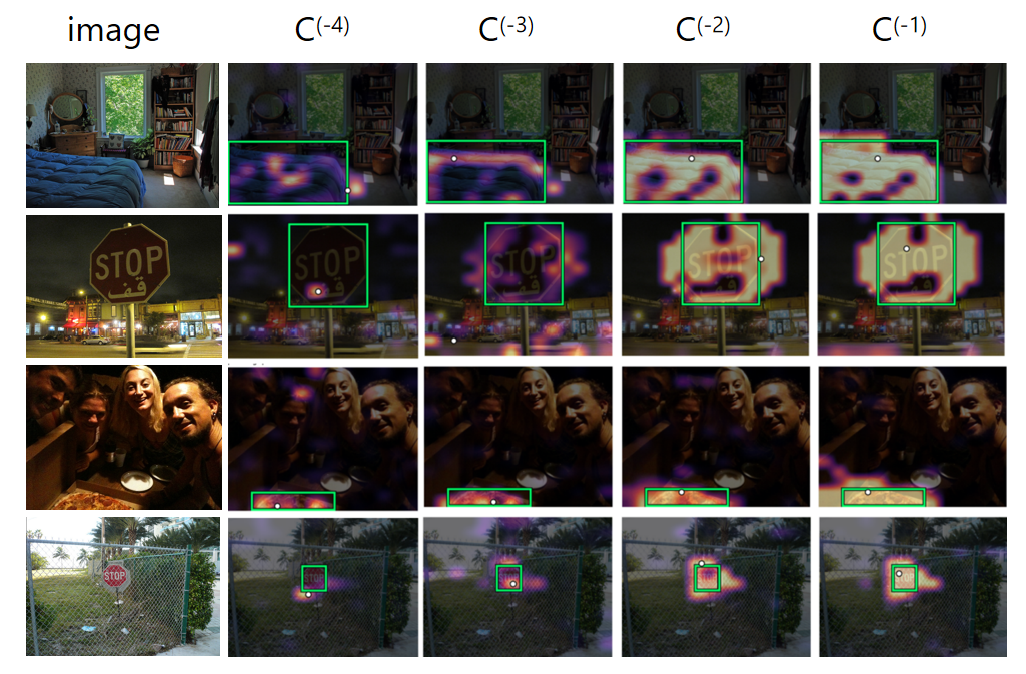}
    \caption{Layer-wise text-conditioned response maps of patch-to-CLS
    contribution features on CLIP ViT-B/16. Green boxes denote ground-truth
    boxes and white dots denote maximum-response patches.}
    \label{fig:layerwise_response}
\end{figure}

To qualitatively examine this layer-wise behavior, we visualize the
text-conditioned responses of contribution features from the last four
visual blocks in Fig.~\ref{fig:layerwise_response}.

\begin{figure*}[t]
\centering
\includegraphics[width=\textwidth]{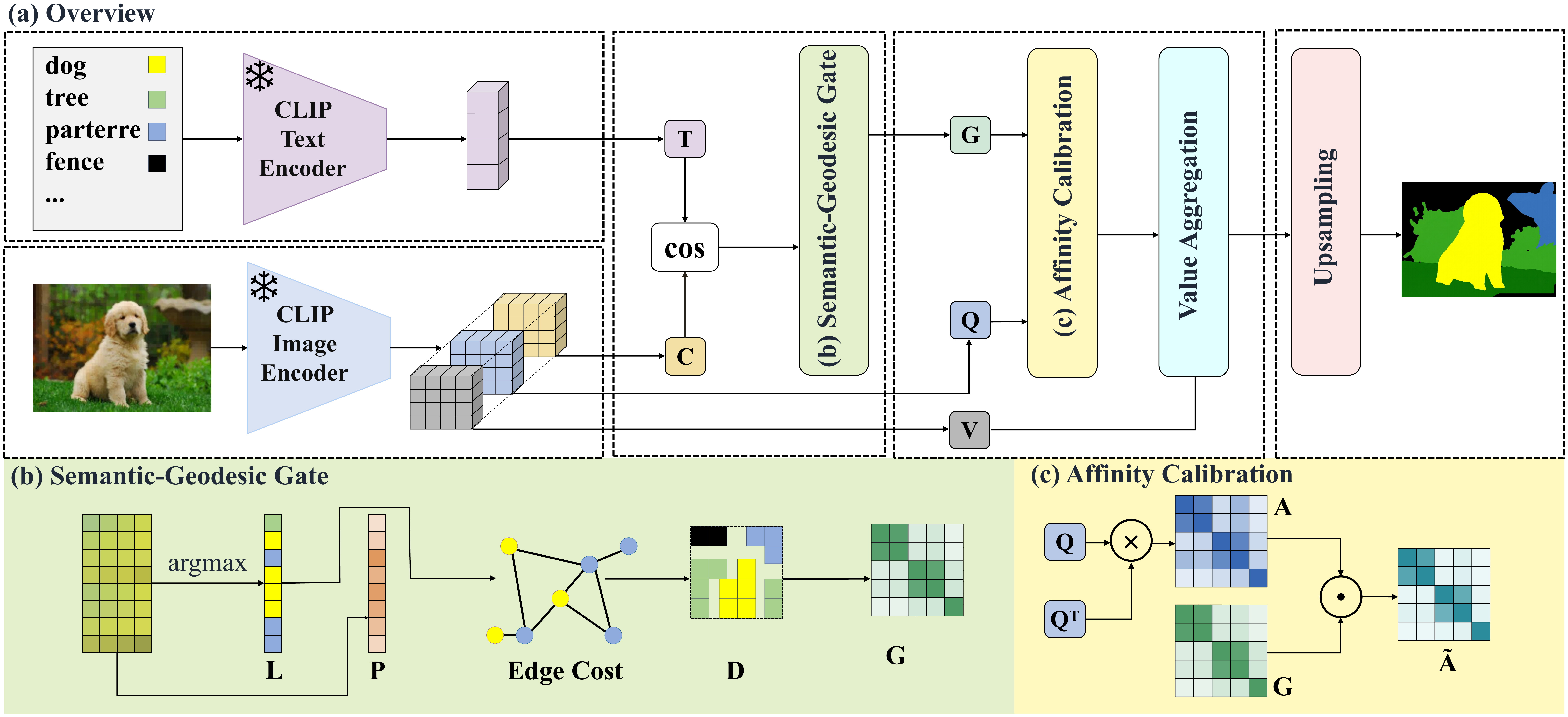}
\caption{
Overview of \textbf{TraceCLIP}.
The frozen CLIP encoders produce text embeddings $T$, patch-to-CLS
contribution features $C$, and final-layer query and value features $Q$
and $V$. Contribution--text similarities yield preliminary semantic
labels $L$ and confidence scores $P$, from which the semantic-geodesic
distance matrix $D$ and topology gate $G$ are constructed. The gate
calibrates the query-query affinity $A$ to obtain $\widetilde{A}$, which
aggregates $V$ to reconstruct dense patch features for final segmentation.
Head indices are omitted for clarity. The exponent $\lambda$, numerical
stabilizer $\varepsilon$, and row normalization in affinity calibration
are also omitted from the diagram for visual clarity.
}
\label{fig:framework}
\end{figure*}

Figure~\ref{fig:layerwise_response} confirms the same trend. Earlier
contribution features produce diffuse responses, whereas near-final features
become more concentrated on the target object, with maximum-response patches
moving closer to annotated regions from $C^{(-4)}$ to $C^{(-1)}$.

\paragraph{Late-block component analysis.}
We next inspect the last transformer block to identify where the gap between
contribution features and final patch states emerges. Let $X_{\mathrm{in}}$
denote the block input, $U$ the self-attention output, and
$X_{\mathrm{attn}}=X_{\mathrm{in}}+U$ the attention-updated state. Let $M$
denote the MLP output and
$X_{\mathrm{out}}=X_{\mathrm{attn}}+M$ the final block output.
After CLIP's post-layer normalization and visual projection,
$X_{\mathrm{out}}$ corresponds to the \emph{Final patch} readout.

We evaluate the block input, attention output, attention-updated state, MLP
output, final output, and patch-to-CLS contribution feature separately.
We also use a mean-MLP replacement, in which the MLP output at every patch
position is replaced by its spatial mean before residual addition. This
removes patch-specific MLP variation while preserving the shared update,
allowing us to test whether the degradation is primarily caused by the MLP.

\begin{table}[!t]
\centering
\small
\begin{tabular}{lccc}
\toprule
\textbf{Source}
& \textbf{Reg. Acc.}
& \textbf{Point. Acc.}
& \textbf{In--Out Gap} \\
\midrule

\multicolumn{4}{c}{\textit{ViT-B/16}} \\
\midrule
Input
& 5.2
& 15.7
& -0.013 \\

Attn. output
& 35.7
& 48.9
& 0.091 \\

Attn. + residual
& 15.1
& 19.2
& -0.043 \\

MLP output
& 0.4
& 26.2
& 0.002 \\

Final output
& 11.3
& 19.9
& -0.042 \\

Mean-MLP repl.
& 10.6
& 18.1
& -0.046 \\

Contribution
& \textbf{43.1}
& \textbf{65.7}
& \textbf{0.275} \\

\midrule
\multicolumn{4}{c}{\textit{ViT-L/14}} \\
\midrule
Input
& 0.9
& 12.2
& -0.014 \\

Attn. output
& 37.9
& 55.5
& 0.111 \\

Attn. + residual
& 9.9
& 13.8
& -0.048 \\

MLP output
& 3.2
& 26.4
& 0.013 \\

Final output
& 8.7
& 14.0
& -0.048 \\

Mean-MLP repl.
& 8.7
& 13.6
& -0.050 \\

Contribution
& \textbf{47.1}
& \textbf{68.9}
& \textbf{0.287} \\

\bottomrule
\end{tabular}

\caption{Late-block component analysis on the COCO validation set.
Reg. Acc. and Point. Acc. are reported in percentages.
In--Out Gap denotes the inside--outside target-category response
difference.}
\label{tab:late_block_analysis}
\end{table}

Table~\ref{tab:late_block_analysis} shows that the final self-attention
block actively produces useful local semantic evidence. On ViT-B/16, the
attention output improves Region Acc. from 5.2\% to 35.7\% and Pointing Acc.
from 15.7\% to 48.9\% relative to the block input, with a similar trend on
ViT-L/14.

However, much of this advantage disappears after residual addition:
Region Acc. drops from 35.7\% to 15.1\% on ViT-B/16 and from 37.9\% to
9.9\% on ViT-L/14. The weak MLP output and mean-MLP replacement indicate
that the degradation is not introduced solely by the final MLP. Instead,
local semantics become less directly accessible when the attention update
is mixed with the residual patch states.

Patch-to-CLS contributions achieve the strongest results, outperforming the complete attention output. This indicates that CLS attention
weighting further selects the patch components most relevant to the
language-aligned global representation. Thus, local semantics are present
in the frozen encoder but are exposed more clearly in the contribution
pathway than through the final patch-token interface. This observation
motivates TraceCLIP to use contribution features as semantic guidance while
retaining final-layer affinity and value features for dense reconstruction.

\section{TraceCLIP}

The previous section shows that patch-to-CLS contribution features expose
more reliable local semantic evidence than final patch states. Directly
using them as segmentation scores, however, ignores the spatial aggregation
required by dense prediction. TraceCLIP therefore converts contribution
responses into a semantic-geodesic topology gate that calibrates final-layer
patch affinity. The entire framework is training-free and keeps both CLIP
encoders frozen, as illustrated in Fig.~\ref{fig:framework}.

\subsection{Contribution-Based Semantic Prior}

Given an image and candidate category names, the frozen CLIP encoders
produce normalized text embeddings and final-layer patch-to-CLS
contribution features, which lie in the shared image-text embedding space
and can therefore be directly compared.

Let $C\in\mathbb{R}^{N\times d}$ denote the contribution feature matrix,
where $N$ is the number of patches and $d$ is the joint embedding
dimension. Let $T\in\mathbb{R}^{K\times d}$ denote the text embedding
matrix of $K$ candidate classes. The contribution-based response matrix is
\begin{equation}
S^{\mathrm{con}}
=
C T^{\mathsf{T}}.
\label{eq:contribution_score}
\end{equation}
For each patch $i$, we take the class with the largest response as its
preliminary semantic label $L_i$. The corresponding softmax probability,
computed after applying the logit scale, is used as its confidence score
$P_i$. These preliminary labels and confidence scores provide semantic region and boundary
cues for subsequent affinity calibration.

\subsection{Semantic-Geodesic Topology Gate}

Final-layer patch affinity can aggregate spatial information, but it may
also propagate features across semantic boundaries. To restrict such
propagation, we build a graph over the patch grid using the preliminary
labels $L$ and confidence scores $P$. Each patch is a node, and edges are
built only between four-neighboring patches.

For a neighboring patch pair $(i,j)$, let $m_{ij}=1$ if
$L_i\neq L_j$ and $m_{ij}=0$ otherwise. We define the boundary confidence
as $c_{ij}=\min(P_i,P_j)$. The edge cost is
\begin{equation}
w_{ij}
=
1+\beta m_{ij}c_{ij},
\label{eq:edge_cost}
\end{equation}
where $\beta$ controls the semantic boundary penalty. Same-label neighbors
have the base cost $1$, while different-label neighbors receive an
additional confidence-weighted cost. Thus, confident semantic boundaries
become harder to cross.

We compute the shortest-path distance $D_{ij}$ between every pair of
patches on this graph and collect the distances into
$D\in\mathbb{R}^{N\times N}$. This distance measures how easily
information can travel from patch $i$ to patch $j$ through four-neighbor
connections. We then convert the distance into a topology gate:
\begin{equation}
G_{ij}
=
\frac{
\exp\left(-D_{ij}^{2}/2\sigma^{2}\right)
}{
\sum_{r=1}^{N}
\exp\left(-D_{ir}^{2}/2\sigma^{2}\right)
}.
\label{eq:semantic_gate}
\end{equation}
The resulting matrix $G\in\mathbb{R}^{N\times N}$ is the
semantic-geodesic topology gate. The parameter $\sigma$ controls its
effective geodesic range: a larger $\sigma$ preserves stronger connections
over longer geodesic distances, whereas a smaller $\sigma$ concentrates
the gate on geodesically closer patches. Each row of $G$ sums to one.

\subsection{Affinity Calibration and Dense Reconstruction}

We next use the semantic-geodesic topology gate to calibrate final-layer
patch affinity. Let $Q^h$ and $V^h$ denote the final-layer query and value
feature matrices in attention head $h$, respectively, and let $d_h$ be the
head dimension. Following ClearCLIP~\cite{lan2024clearclip}, we construct
the patch affinity using query-query similarity:
\begin{equation}
A^h
=
\operatorname{softmax}_{\mathrm{row}}
\left(
\frac{
Q^h(Q^h)^{\mathsf{T}}
}{
\sqrt{d_h}
}
\right).
\label{eq:qq_affinity}
\end{equation}
Here $A^h$ is the final-layer patch affinity matrix in head $h$.
Query-query similarity provides an effective measure of patch-level
correlation for dense CLIP inference, but it may still connect visually
similar patches across semantic boundaries.

We therefore calibrate $A^h$ using the contribution-derived topology gate:
\begin{equation}
\widetilde{A}_{ij}^{h}
=
\frac{
(A_{ij}^{h}+\varepsilon)
(G_{ij}+\varepsilon)^{\lambda}
}{
\sum_{r=1}^{N}
(A_{ir}^{h}+\varepsilon)
(G_{ir}+\varepsilon)^{\lambda}
}.
\label{eq:affinity_calibration}
\end{equation}
Here $\lambda$ controls the calibration strength and $\varepsilon$ is a
small constant for numerical stability. A patch pair receives a high
propagation weight only when it is supported by both final-layer visual
affinity and contribution-derived semantic topology. Consequently,
cross-boundary propagation is suppressed even when the visual affinity
between the patches is high.

Finally, the calibrated affinity $\widetilde{A}^h$ aggregates the
final-layer value features $V^h$ in each attention head. The resulting
head features are concatenated, passed through the final attention output
projection, and then processed by the CLIP visual post-layer normalization
and visual projection. The reconstructed patch features are normalized in
the image-text embedding space and compared with class text embeddings to
obtain dense class scores. The patch-level score maps are upsampled to the
image resolution as the final prediction.
\begin{table*}[!t]
\centering
\footnotesize
\setlength{\tabcolsep}{1.2pt}

\begin{tabular}{@{}lcccccccccccc@{}}
\toprule
&
&
&
\multicolumn{6}{c}{\textbf{Without Background}}
&
\multicolumn{4}{c}{\textbf{With Background}} \\
\cmidrule(lr){4-9}
\cmidrule(lr){10-13}

\textbf{Method}
& \textbf{Training-free}
& \textbf{Encoder}
& \textbf{VOC20}
& \textbf{Ctx59}
& \textbf{Stuff}
& \textbf{City.}
& \textbf{ADE20K}
& \textbf{Avg.}
& \textbf{VOC21}
& \textbf{Ctx60}
& \textbf{Object}
& \textbf{Avg.} \\
\midrule

GroupViT
& No
& ViT-S/16
& 79.7
& 23.4
& 15.3
& 11.1
& 9.2
& 27.7
& 50.4
& 18.7
& 27.5
& 32.2 \\

TCL
& No
& ViT-B/16
& 77.5
& 30.3
& 19.6
& 23.1
& 14.9
& 33.1
& 51.2
& 24.3
& 30.4
& 35.3 \\

\midrule

CLIP
& Yes
& ViT-B/16
& 41.8
& 9.2
& 4.4
& 5.5
& 2.1
& 12.6
& 16.2
& 7.7
& 5.5
& 9.8 \\

MaskCLIP
& Yes
& ViT-B/16
& 74.9
& 26.4
& 16.4
& 12.6
& 9.8
& 28.0
& 38.8
& 23.6
& 20.6
& 27.7 \\

ReCo
& Yes
& ViT-B/16
& 57.7
& 22.3
& 14.8
& 21.1
& 11.2
& 25.4
& 25.1
& 19.9
& 15.7
& 20.2 \\

SCLIP
& Yes
& ViT-B/16
& 80.4
& 34.2
& 22.4
& 32.2
& 16.1
& 37.1
& \textbf{59.1}
& 30.4
& 30.5
& 40.0 \\

ClearCLIP
& Yes
& ViT-B/16
& 80.9
& 35.9
& 23.9
& 30.0
& 16.7
& 37.5
& 51.8
& 32.6
& 33.0
& 39.1 \\

NACLIP
& Yes
& ViT-B/16
& 79.7
& 35.2
& 23.3
& 35.5
& 17.4
& 38.2
& 58.9
& 32.2
& 33.2
& 41.4 \\

TraceCLIP
& Yes
& ViT-B/16
& \textbf{83.1}
& \textbf{36.6}
& \textbf{24.6}
& \textbf{37.0}
& \textbf{17.5}
& \textbf{39.8}
& 58.8
& \textbf{34.7}
& \textbf{34.5}
& \textbf{42.7} \\

\midrule

CLIP
& Yes
& ViT-L/14
& 15.8
& 4.5
& 2.4
& 2.9
& 1.2
& 5.4
& --
& --
& --
& -- \\

MaskCLIP
& Yes
& ViT-L/14
& 30.1
& 12.6
& 8.9
& 10.1
& 6.9
& 13.7
& --
& --
& --
& -- \\

SCLIP
& Yes
& ViT-L/14
& 60.3
& 20.5
& 13.1
& 17.0
& 7.1
& 23.6
& 44.4
& 22.3
& 24.9
& 30.5 \\

ClearCLIP
& Yes
& ViT-L/14
& 80.0
& 29.6
& 19.9
& 27.9
& 15.0
& 34.5
& 48.7
& 28.3
& 29.7
& 35.5 \\

TraceCLIP
& Yes
& ViT-L/14
& \textbf{85.0}
& \textbf{33.1}
& \textbf{22.6}
& \textbf{36.1}
& \textbf{18.1}
& \textbf{39.0}
& \textbf{51.5}
& \textbf{31.1}
& \textbf{32.7}
& \textbf{38.4} \\

\bottomrule
\end{tabular}

\caption{Zero-shot semantic segmentation results in mIoU on datasets
without and with a background category. The two Avg. columns denote the
means over VOC20, Context59, COCO-Stuff, Cityscapes, and ADE20K, and over
VOC21, Context60, and COCO-Object, respectively. Best results among
training-free methods are shown in bold. Dashes indicate results not reported by the
corresponding source.}
\label{tab:main_results}
\end{table*}

\section{Experiments}

\subsection{Experimental Setup}

\paragraph{Datasets and metrics.}
We evaluate zero-shot semantic segmentation on eight standard benchmarks.
The no-background setting includes VOC20~\cite{everingham2010pascal},
Context59~\cite{mottaghi2014role}, COCO-Stuff~\cite{caesar2018coco},
Cityscapes~\cite{cordts2016cityscapes}, and ADE20K~\cite{zhou2019semantic};
the with-background setting includes VOC21~\cite{everingham2010pascal},
Context60~\cite{mottaghi2014role}, and COCO-Object~\cite{lin2014microsoft}.
We report mIoU on all datasets.

\paragraph{Backbones and prompts.}
We evaluate frozen CLIP ViT-B/16 and ViT-L/14 without updating either
encoder. For each class, we average normalized text features over the
standard OpenAI ImageNet prompt templates. When a category contains multiple
synonyms, their logits are merged into the corresponding semantic class
during post-processing.

\paragraph{Implementation details.}
We use sliding-window inference for high-resolution dense prediction.
For all datasets except Cityscapes, images are resized with aspect ratio
preserved under a maximum resolution of $2048\times448$; Cityscapes is
evaluated at its original resolution. We use a $448\times448$ window with
stride $224$. Patch logits are bilinearly upsampled to the window resolution,
averaged over overlapping windows, and finally upsampled to the original
image size. The logit scale $\gamma$ is set to 50. No training, test-time
optimization, or stochastic sampling is used.

We use the same hyperparameter protocol for all datasets and backbones.
We set $\beta=1.0$ and $\lambda=1.5$. For each dataset, $\sigma$ is fixed
throughout evaluation, with a scale coefficient of $0.4$ used to determine
its value. We use $\varepsilon=10^{-6}$ as a standard numerical stabilizer.
These settings remain unchanged across all experiments and ablations.

\paragraph{Baselines.}
We compare with GroupViT~\cite{xu2022groupvit},
TCL~\cite{cha2023learning}, CLIP~\cite{radford2021learning},
MaskCLIP~\cite{zhou2022extract}, ReCo~\cite{shin2022reco},
SCLIP~\cite{wang2024sclip}, ClearCLIP~\cite{lan2024clearclip},
and NACLIP~\cite{hajimiri2025pay} under the same dataset protocol.
The reported scores for all baselines except NACLIP follow
ClearCLIP~\cite{lan2024clearclip}, while NACLIP results are taken
from~\cite{hajimiri2025pay} under the comparable no-post-processing
setting. All TraceCLIP results are obtained from our own implementation
without training or external vision models.

\subsection{Main Results}

Table~\ref{tab:main_results} reports zero-shot semantic segmentation
results with and without a background category. With ViT-B/16,
TraceCLIP achieves 39.8 and 42.7 average mIoU in the two settings,
outperforming the strongest prior training-free baselines by 1.6 and
1.3 points, respectively. With ViT-L/14, TraceCLIP improves the
corresponding averages by 4.5 and 2.9 points. These results show that
contribution-derived semantic topology consistently improves average
performance across both backbones and evaluation settings while keeping
CLIP fully frozen.

\subsection{Analysis and Ablation}
\begin{table}[!t]
\centering
\footnotesize

\textbf{(a) Dense readout analysis}

\smallskip

\setlength{\tabcolsep}{1.3pt}
\begin{tabular}{@{}lcccccc@{}}
\toprule
Source & VOC20 & Ctx59 & Stuff & City. & ADE20K & Avg. \\
\midrule
\emph{Final patch}
& 38.1
& 8.8
& 4.2
& 6.9
& 1.7
& 11.9 \\

$C^{(-3)}$
& 33.1
& 16.6
& 8.8
& 16.0
& 8.0
& 16.5 \\

$C^{(-2)}$
& 53.4
& 17.4
& 11.3
& 13.0
& 6.9
& 20.4 \\

$C^{(-1)}$
& \textbf{57.0}
& \textbf{22.9}
& \textbf{14.7}
& \textbf{22.6}
& \textbf{10.6}
& \textbf{25.6} \\
\bottomrule
\end{tabular}

\medskip

\textbf{(b) Component ablation}

\smallskip

\setlength{\tabcolsep}{2.4pt}
\begin{tabular}{@{}lccc@{}}
\toprule
Variant & Weight & Aggregate $V$ & Avg. \\
\midrule
Direct $V^{(-1)}$ readout
& None
& No
& 26.3 \\

$A^h$-weighted $V$
& $A^h$
& Yes
& 37.5 \\

$G$-weighted $V$
& $G$
& Yes
& 30.1 \\

TraceCLIP
& $\widetilde{A}^h$
& Yes
& \textbf{39.8} \\
\bottomrule
\end{tabular}

\caption{Analysis and ablation results with CLIP ViT-B/16 on datasets
without a background category, reported in mIoU. In (a), Avg. denotes
the mean over VOC20, Context59, COCO-Stuff, Cityscapes, and ADE20K.
In (b), all variants are implemented and evaluated under the same
inference pipeline. Best results in each subtable are shown in bold.}
\label{tab:analysis_ablation}
\end{table}

\paragraph{Dense readout analysis.}
Table~\ref{tab:analysis_ablation}(a) evaluates patch-to-CLS
contributions as dense semantic readouts. Final patch states achieve only
11.9 average mIoU, whereas $C^{(-3)}$, $C^{(-2)}$, and $C^{(-1)}$
obtain 16.5, 20.4, and 25.6, respectively. Although the trend is not
strictly monotonic on every dataset, $C^{(-1)}$ performs best across all
five benchmarks and more than doubles the average performance of final
patch states. This confirms that near-final contributions expose stronger
local semantics. However, their direct-readout result remains below the
39.8 of TraceCLIP, indicating that effective dense prediction also requires
spatial feature aggregation.

\paragraph{Component ablation.}
Table~\ref{tab:analysis_ablation}(b) examines the roles of visual affinity
and semantic topology. Directly reading $V^{(-1)}$ achieves 26.3 average
mIoU, while aggregation with query-query affinity $A^h$ improves it to
37.5, showing the importance of visual feature propagation. Using the
semantic-geodesic gate $G$ alone obtains 30.1, suggesting that it provides
useful region and boundary cues but cannot replace visual affinity. The
full TraceCLIP achieves 39.8, outperforming $A^h$ alone by 2.3 points.
This verifies that contribution-derived topology complements visual
affinity by suppressing propagation across semantic boundaries.

\section{Conclusion}
We investigate where local vision-language semantics become most explicitly
accessible within frozen CLIP. Our diagnostic results show that patch-to-CLS
contribution features provide substantially stronger category-aware and
spatially grounded evidence than final patch-token representations. The
late-block analysis further indicates that such evidence is clearly exposed
in the attention contribution pathway but becomes less directly accessible after residual mixing.
Based on this observation, we introduce TraceCLIP, which converts contribution-derived semantic responses into a semantic-geodesic topology gate for calibrating final-layer patch affinity and reconstructing dense features. Across eight zero-shot semantic segmentation
benchmarks, TraceCLIP improves average mIoU over the strongest prior
training-free methods by 1.3--4.5 points across both backbones and background
settings, without additional training, external vision foundation models, or
region-level supervision.
These findings point to a broader representational phenomenon:
task-relevant local semantics may remain encoded in the internal
construction of a globally aligned representation. Tracing such contribution
pathways therefore provides a useful perspective on how semantic information
is formed, selected, and obscured within frozen vision-language models.
\bibliographystyle{plainnat}
\bibliography{references}

\end{document}